# Handling missing values in healthcare data: A systematic review of deep learning-based imputation techniques

Mingxuan Liu[1#], Siqi Li[1#], Han Yuan[1], Marcus Eng Hock Ong[2,3], Yilin Ning[1], Feng Xie[1,2], Seyed Ehsan Saffari[1,2], Victor Volovici[4], Bibhas Chakraborty[1,2,5,6], Nan Liu[1,2,7,8*]

[1] Centre for Quantitative Medicine, Duke-NUS Medical School, Singapore

[2] Programme in Health Services and Systems Research, Duke-NUS Medical School, Singapore

[3] Department of Emergency Medicine, Singapore General Hospital, Singapore

[4] Department of Neurosurgery, Erasmus MC University Medical Center, Rotterdam, The Netherlands

[5] Department of Statistics and Data Science, National University of Singapore, Singapore

[6] Department of Biostatistics and Bioinformatics, Duke University, Durham, NC, USA

[7] SingHealth AI Office, Singapore Health Services, Singapore

[8] Institute of Data Science, National University of Singapore, Singapore

[#] These authors contributed equally

[*] Corresponding author: Nan Liu, Centre for Quantitative Medicine, Duke-NUS Medical School, 8 College Road, Singapore 169857, Singapore. Phone: +65 6601 6503. Email: liu.nan@duke-nus.edu.sg

## Abstract

### Objective

The proper handling of missing values is critical to delivering reliable estimates and decisions, especially in high-stakes fields such as clinical research. The increasing diversity and complexity of data have led many researchers to develop deep learning (DL)-based imputation techniques. We conducted a systematic review to evaluate the use of these techniques, with a particular focus on data types, aiming to assist healthcare researchers from various disciplines in dealing with missing values.

**Materials and Methods**

We searched five databases (MEDLINE, Web of Science, Embase, CINAHL, and Scopus) for articles published prior to August 2021 that applied DL-based models to imputation. We assessed selected publications from four perspectives: health data types, model backbone (i.e., main architecture), imputation strategies, and comparison with non-DL-based methods. Based on data types, we created an evidence map to illustrate the adoption of DL models.

**Results**

We included 64 articles, of which tabular static (26.6%, 17/64) and temporal data (37.5%, 24/64) were the most frequently investigated. We found that model backbone(s) differed among data types as well as the imputation strategy. The "integrated" strategy, that is, the imputation task being solved concurrently with downstream tasks, was popular for tabular temporal (50%, 12/24) and multi-modal data (71.4%, 5/7), but limited for other data types. Moreover, DL-based imputation methods yielded better imputation accuracy in most studies, compared with non-DL-based methods.

**Conclusion**

DL-based imputation models can be customized based on data type, addressing the corresponding missing patterns, and its associated "integrated" strategy can enhance the efficacy of imputation, especially in scenarios where data is complex. Future research may focus on the portability and fairness of DL-based models for healthcare data imputation.

**Abbreviations**: DL, Deep Learning; AE, Auto Encoder; DAE, Deep Auto Encoder; GAN, Generative Adversarial Network; LSTM, Long Short-term Memory; MLP, Multilayer Perceptron; RNN, Recurrent Neural Networks

# 1. Introduction

Healthcare data has emerged in diverse formats in the era of big data. Personalized health monitoring devices, for instance, enable the collection of data tailored to an individual's activities on a daily basis. Likewise, rapidly evolving laboratory techniques generate vast amounts of sequencing data. However, these new data formats are more susceptible to the problem of missing values than traditional tabular clinical data collected in a prospective observational or randomized trial.

Missing values cast a shadow on data analysis: they can reduce prediction power and lead to bias in downstream decision-making[1, 2], which is particularly problematic in high-fidelity decision-making situations, such as those in healthcare. Simple complete data analysis or simple imputation (mean, median, or mode) may resolve missingness for tabular static data, but such strategies may not be suitable for a variety of data types and architectures, ranging from static to temporal, tabular to imaging and sequencing data. As a result, advanced approaches are necessary to ensure model quality and robustness.

As described by Little and Rubin[3, 4], missing data can be categorized into three types: missing completely at random (MCAR), missing at random (MAR), and missing not at random (MNAR). Prior to the widespread adoption of deep learning (DL), traditional statistical and machine learning approaches, such as interpolation methods, k-nearest neighbor (k-NN)[5], multiple imputation by chained equations (MICE)[6], random forest (RF)-based models like MissForest[7], have been utilized to impute missing values. However, these widely used methods may be limited to certain categories of missing data; for instance, MICE generally requires the assumption that the data missing type is MAR[6]. When applied to complex healthcare data, these non-DL-based imputation strategies may also suffer from low accuracy[8, 9], especially when the mechanism of missingness is uncertain.

In recent years, DL-based methods have been used increasingly for solving missing value problems and demonstrated their ability in enhancing imputation accuracy[10, 11]. Furthermore, DL-based models can be tailored to handle complex missing patterns and data structures, such as time-series data with unique sequential structures and image data with spatial patterns[12, 13]. With superior performance and designation flexibility, DL-based imputation models have become popular in a variety of applications, from in-patient mortality prediction to early detection of Alzheimer's Disease (AD).

Although there are several existing reviews on missing value imputation, most of them either focus on non-DL-based methods[14, 15], or treat the neural network as a single type of method[16, 17]. Due to the lack of specificity, these articles cannot adequately assist prospective researchers who contemplate the application of DL-based models to their own data. To our knowledge, there has not been a systematic review of DL-based missing value imputation methods for diverse types of healthcare data. To address this gap, we present an evidence map analysis[18] that examines model-use by data type, and we aim to provide guidance for managing missing values with DL-based methodologies for researchers engaged in different clinical disciplines.

# 2. Materials and Methods

## 2.1 Search Strategies

In this study, we undertook a systematic literature search to identify relevant research articles. We searched five databases (MEDLINE, Web of Science, Embase, CINAHL, and Scopus) using a combination of search phrases "missing value", "imputation", "machine learning", "deep learning", and "healthcare". Detailed search strategies are provided in eTable 1.

## 2.2 Exclusion Criteria

We conducted the study according to the Preferred Reporting Items for Systematic Reviews (PRISMA) guidelines[19]. Papers with any of the following reasons were excluded: the study was not in the medical or clinical domain, the imputation model used was not DL-based or not specified, the study was not published as a research article (e.g., conference poster, conference abstract, or book chapter), or the article was not written in English.

## 2.3 Selection Procedure and Data Extraction

Two reviewers (ML and SL) independently screened the titles and abstracts based on the eligibility criteria between 6 August and 11 September 2021. Discrepancies were resolved through discussions with a third reviewer (HY). The full-text screening and information extraction took place from 12 September to 22 October 2021, during which ML and SL separately accessed the documents and, in the event of disagreement, consulted with HY. Four aspects of information were extracted from the included articles: data types, model backbones (i.e., main architectures), imputation strategies, and comparison with non-DL-based methods.

## 2.4 Data Analysis

To generate an evidence map[18] that illustrates the application of DL-based imputation models across various data types, we classified the types of data involved in imputation into six categories: tabular static data, tabular temporal data, genetic and genomic data, image data, signal data, and multi-modal data. Tabular static data and tabular temporal data both contain observations as rows and features as columns, but only tabular temporal data include the time factor. Genetic and genomic data encompasses both DNA data for organisms and personal genetic information. Image data and signal data refer to the information generated by specific medical devices, such as magnetic resonance imaging (MRI) and electrocardiogram (ECG). Multi-modal data refers to the use of multiple data types in a single imputation task.

Next, we categorized the articles according to the "backbones" of the imputation model: 1) multi-layer perceptron (MLP); 2) recurrent neural network (RNN), including vanilla RNN, long short-term memory (LSTM), and gated recurrent unit (GRU); 3) the framework of autoencoder (AE) which includes vanilla autoencoder, denoising autoencoder (DAE), and variational autoencoder (VAE); 4) the framework of generative adversarial network (GAN); 5) a hybrid of the four backbones mentioned above; and 6) other less common models, such as self-organizing map (SOM). The precise definitions of these models and their accompanying general imputation mechanisms are described in Table 1. To assess the imputation strategy, we divided it into two categories: separated and integrated. "Separated" means that the imputation process is separated from downstream tasks such as disease classification and risk prediction, whereas "integrated", also known as "end-to-end", means that the imputation process is addressed concurrently with downstream tasks.

The evidence map was presented based on the cross-tabulation of model backbones and data types. In addition, a bar plot displaying the distribution of imputation strategies was provided. Python version 3.8.3 (Python Software Foundation, Delaware, USA) and R version 4.0.2 (The R Foundation for Statistical Computing) were used for data analysis.

## 3. Results

Our search of five databases yielded 1058 papers, of which 64 were included for analysis. The details of the selection procedure are illustrated in Figure 1. Figure 2 depicts the evidence map between the "backbones" (i.e., main architectures) of DL-based missing value imputation models and the types of healthcare data. Among the 64 studies, 17 presented missing value imputation models for tabular static data, 24 for tabular temporal data, seven for genetic and genomic data, five for image data, five for signal data, and seven for multi-modal data. These numbers are not mutually exclusive as a single study may involve more than one data type. MLP model for tabular static data and vanilla autoencoder for tabular temporal data are the most common (9.4%, 6/64) among all cells in the cross-tabulation.

As indicated in Figure 3, most studies (70%, 45/64) adopted the "separated" strategy. The "integrated" strategy was popular among tabular temporal data and multi-modal data, but less used for tabular static data and genetic and genomic data, and rarely applied to image or signal data. Moreover, 35 of the 64 selected papers investigated the type of data missingness, 18 of which investigated MNAR missingness, while the rest developed imputation methods directly without examining this specification. Next, we present DL-based imputation techniques based on various types of health data.

## 3.1 Tabular Static Data

A total of 17 studies[20-36] used tabular static data in this review. Most (76%, 13/17) of them provided non-DL-based imputation methods as baselines for comparison with DL-based methods in terms of imputation accuracy (if complete data is available) and prediction performance, with simple imputation (mean/median/mode imputation, 41%, 7/17), RF-based methods (35%, 6/17), MICE (36%, 6/17), and k-NN (29%, 5/17) being the most common options. Most of these studies demonstrated the superiority of DL-based approaches.

The structures of the DL-based imputation methods for tabular static data are dominated by MLP-oriented models. Six studies used MLP models directly[23, 24, 30-32, 36], the other five created autoencoder models employing MLP modules as encoders and decoders[22, 25-27, 33], and three studies that utilized GANs also implemented simple MLP modules as generators and discriminators. In addition to MLP, this review also included several alternative options. Dynamic layered RNN[28] was applied to repeat the imputation process with an aim to enhance the imputation accuracy, an autoencoder with a structure of ladder network was developed to improve the imputation robustness in terms of different missing rates[34], and SOM was used in a comparison study which indicated the superiority of SOM in terms of improvement downstream task (classification) performance[29].

The handling of mixed variable types, such as numerical, categorical, and ordinal data, is a challenge for tabular data (both static and temporal), which is due to the assumptions about data distribution and limitations of some non-DL-based methods[26]. However, DL-based imputation methods are able to accommodate mixed variable types with proper encoding and activation functions[25].

## 3.2 Tabular Temporal Data

Among the 64 included studies, 24 [8, 9, 37-58] addressed missing value imputation for tabular temporal data. Missingness for this type of data is typically caused by factors that are less controllable, such as the dropping out of patients or the use of different assessment patterns for different patient subgroups[37]. Therefore, informative missingness with time dynamics and high missing rates pose challenges to imputation.

RNN-based (46%, 11/24) and AE-based (38%, 9/24) methods were commonly employed to impute tabular temporal data, where the "separated" imputation strategy was more often used in conjunction with the latter (27%, 3/11 for RNN-based versus 78%, 7/9 for AE-based). The development of the "separated" imputation strategy relied on the ground truth, that is, artificially composing missingness by dropping values at random and comparing the imputed values to the dropped ground truth. Among 64 studies, seven studies composed such missingness with various rates ranging from 5% to 90% in order to assess the robustness of imputation against different missing rates. Six (67%, 6/9) of the studies that applied AE-based methods artificially composed missing values and adopted the “separated” imputation strategy.

The nature of modeling sequence data makes RNN-based methods capable of capturing time dynamics-related missing patterns. Among these 11 studies that applied RNN-based methods, three studies[8, 49, 57] designed the models with bi (multi)-directions to utilize both past and future information for imputation, while another three[38, 40, 43] applied GRU-D – a model equipped with a specific parameter to characterize the decay of effects as time passes – or developed variants of GRU-D. In addition, two studies developed hybrid RNN-based and AE-based methods in which the AE component was added following the initial imputation by the RNN component[39, 46]. Another two studies utilized the framework of GAN, where adversarial learning (alone[9] or in combination with RNN[45]) can contribute to avoiding error propagation from imputation to downstream tasks.

## 3.3 Genetic and Genomic Data

Seven studies[59-65] dealt with genetic and genomic data, including sequencing data[60, 62, 63, 65] and microarray data[59, 61, 64]. Sequencing data may comprise around 50% zero count observations[62, 65], some of which are "false" zeros, i.e., missing values due to insufficient sequencing input[60]. Furthermore, the high dimension property presents a barrier to the imputation of sequencing data. In microarray data, 90% of genes may have one or more missing values, as a consequence of experimental factors in the process, such as hybridization failures, artifacts on the microarray, insufficient resolution, image noise, and corruption, which are difficult to resolve[66].

Four studies[61-63, 65] used AE-based models (vanilla AE and VAE), and the remaining studies[59, 60, 64] employed MLP, where biological background knowledge may be incorporated into the imputation process for both types of models. With AE-based models, Badsha et al.[62] used transfer learning to extract prior knowledge about gene-gene relationships and learn the dependence structure within the reference panel. According to Kinalis et al.[63], the architecture of AE and VAE allows for more interpretable imputation procedures. With appropriate training, the autoencoder can be interpreted as a combination of biologically meaningful modules. Apart from AE's natural ability to reduce the dimension in the latent space, MLP could also contribute to the development of the dimension deduction approach for imputation[59, 60]. As an example, Chen et al.[59] deciphered two low-dimensional hidden representations from the original high-dimensional data to explain a molecular relationship and a sample level connection, respectively.

## 3.4 Image Data

Among the five papers[12, 67-70] that focused on image data, five targeted neuroimages, including magnetic resonance imaging (MRI) and positron emission tomography (PET) that are widely used for computer-aided diagnosis of AD and mild cognitive impairment (MCI)[12, 68-70]. Another study[67] dealt with cardiac magnetic resonance (CMR) images, which are often considered the gold standard for many cardiovascular medicine analyses. Insufficient image quality and acquisition or storage errors are common causes of pixel-level missingness in image data[12, 67]. In practice, MRI patients may reject PET scans due to their high cost and radioactive exposure[70], resulting in the absence of the entire image.

All five studies designed GAN frameworks for missing value imputation. Only one study[67] established non-DL-based imputation baselines for comparison (mean imputation and an interpolation approach), while the other four did not. Conditional GAN that utilizes label information could be used to impute missing pixels in an image or to synthesize a new one when the whole image is missing[67]. Imputing PET data based on the corresponding MRI data is natural for diagnosing brain diseases such as AD since the usage of multi-source images can provide rich information[12, 71]. In this case, the entire image rather than a few pixel values must be imputed. To impute the entire image, a task-induced GAN can be developed to improve the imputation quality with two tasks designed for the discriminator: whether the image is true (the imputed MRI must be true enough to fool the discriminator) and whether it indicates the disease (the imputed MRI should be consistent with PET)[12]. Cycle-GAN and Colla-GAN have been adopted for this task, which required two discriminators to detect synthetic MRI and PET, respectively[68, 69], similar to the approach proposed by Pan et al. that utilized the connection between MRI and PET and designed two generators and two discriminators in their GAN framework[70].

## 3.5 Signal Data

In this review, various signal data were discussed in five studies, including actigraphy device data[72], smartphone applications data[73], wearable sensors data[74], medical waveform (e.g.

ECG) data[75], time-series measured at each voxel of fMRI[76]. Compared with tabular temporal data, signal data usually has a high sample rate and is susceptible to noise. For data collected in the movement, a common issue is the large amount of missing data with blocks of features concurrently lost as a result of intermittent disconnections, body movement, and firmware malfunction[74], where the missing interval may need to be characterized as a period of continuously repeated zero values[72]. For medical waveform data, even a low missing rate in real-time operational systems, such as removing the lead for a few seconds, can significantly harm the prediction performance over the next hour[75]. Due to the fMRI technique's great susceptibility to noise artifacts, as with image modality, missing values become common for signal modality as well, posing challenges to the imputation process, where the spatiotemporal nature can be of assistance[76].

The non-DL-based imputation method of simple imputation (mean/median imputation, 80%, 4/5) is a common baseline in these five studies. Three studies used AE-based models with customization, such as denoising autoencoder where missingness was treated as a type of noise[72], adversarial autoencoder where the encoder contributed to feature representation and followed by the discriminator of GAN[73], and 1-D convolutional and corresponding deconvolutional modules used by Miller et al. in their autoencoder framework[75]. The other two studies[74, 76] applied RNN models with the time factor into consideration. For signal imputation in fMRI data, the missing values were first filled based on spatial information, and a GRU layer was employed for regularization in the time domain[76].

## 3.6 Multi-modal Data

Seven studies[10, 77-82] investigated imputation methods for multi-modal data, where information fusion procedures were essential in connecting modality-specific models. Four employed autoencoder models and designed linkages between encoders and decoders to concatenate different modalities[10, 77, 78, 82]. For example, Kim et al.[78] developed a stacked DAE model with a merged hidden layer working as the linkage. Additionally, Kim et al.[79] created a collaborative layer to connect MLP models for different modalities. The task-specific layers designed by Thung et al.[80] could enable iterative communication between modality-specific layers, thereby facilitating the exchange of cross-modal information. Vivar et al.[81] established an end-to-end framework, in which the self-attention process aggregated recurrent graph convolutional models. Using this architecture, missing value imputation was transformed into the completion of a geometric matrix. Combining the graph convolutional model with LSTM has shown to be effective in solving this geometric problem.

# 4. Discussion

In this systematic review, we contribute to exhaustively summarizing knowledge regarding the efficacy of deep learning models by investigating their applications to missing value imputation for healthcare data. Having identified popular deep learning models with respect to data type, we found that deep learning models could be customized to take both the data type and the corresponding missing patterns into consideration, thereby improving the quality of missing value imputation. Moreover, the "integrated" imputation strategy could enhance the performance of both imputation and downstream analysis, and the usage varied across data types, highlighting the benefit of imputation based on the data type's characteristics.

Designing and implementing data-type-oriented imputation models could benefit both the imputation process and downstream tasks. The MLP models and autoencoders that use MLP as encoders and decoders are well-suited to capturing the feature relevance of tabular static data in a relatively straightforward manner. When dealing with tabular temporal data, informative missingness and high missing rates are common[38], which makes it difficult to characterize time dynamics. RNN-based models such as GRU-D[38] and its variants are helpful for capturing time patterns and imputing missing values for tabular temporal data. For genetic and genomic data, MLP or AE with MLP acting as encoder and decoder is appropriate for imputation with effectively used biological knowledge.

GAN-based models are commonly used for image imputation. Additional information, such as labeling (e.g., used in Co-GAN[83]) and relevant images (as used in Cycle-GAN[84] or Colla-GAN[69]), can enhance the performance of imputation. Modules from CNN to Transformer[85] to Swin Transformer[86] could be added to the framework to tackle spatial information in image data. The imputation procedure for general signal data may resemble that for tabular temporal data. In multi-model data, the fusion of mode-specific models is indispensable, and most of the current operations are focused at the layer level, for example, stacking and the self-attention mechanism.

There is currently a vacancy in the imputation approach for medical text data. This may be explained by the fact that techniques in natural language processing (e.g., BERT[87]) inherently learn representation through masking, that is, considering some language tokens as missing on purpose, so actual missingness will not be an issue. Healthcare text data can be analyzed using customized biomedical research models, such as BioBERT[88] and MedBERT[89].

Due to the widespread use of diverse and complex data types, the causes and forms of missing data have become more complicated. Instead of imputing data of different types separately, hybrid models are designed to better capture missing patterns and benefit from the interaction among imputation processes through an integral approach[78, 79]. For example, the integration of LSTM and the graphical convolutional model developed by Vivar et al.[81] advanced the performance of imputation and classification in computer-aided diagnosis. Developing hybrid models capable of handling hybrid data will open up possibilities for further research on imputation in complex scenarios. Rather than referring to mixed data types, some researchers use the term "multi-modality" to describe datasets from different sources but of the same type (e.g., MRI and PET images[12, 68]). For this image type of “multi-modality”, GAN-based frameworks that involve image translation[69] and Transformer-based architectures that are intrinsically integrated with attention modules may also be able to solve the imputation problem[85].

Block-building logic enables DL-based models to adopt an “integrated” strategy, i.e., co-training imputation and downstream tasks, which is preferred for several reasons. First, the interaction between these two tasks can be mutually beneficial, reducing the bias in imputation, and providing prior information for the downstream modelling[39, 45, 81]. Second, the “integrated” strategy could be more practical because it circumvents the difficulties in defining imputation accuracy when the missing rate is high, indicating limited ground truth for imputation quality checking[90], or when multiple data types are involved and the overall imputation accuracy is difficult to define. As shown in Figure 3, this is consistent with the relative prevalence of the “integrated” strategy when working with tabular temporal and multi-modal data.

Third, in contrast to the “separated” strategy, the “integrated” strategy does not emphasize the selection of the optimal combination of imputation and downstream models[81]. The step-wise selection employed in the “separated” strategy, i.e., determining the best imputation model first and then deciding on the downstream models, may not be effective given the belief that the accuracy of imputation does not directly affect the accuracy of downstream tasks[24, 81]. Although it is theoretically possible to test all potential combinations of imputation and downstream models and select the optimal combination, doing so would be extremely time-consuming. The “integrated” strategy resolves such practical difficulties by imputing missing data while developing the downstream models. Last, the “integrated” strategy requires little additional effort when both the imputation and downstream models are DL-based, since the parameters can easily be linked up and updated simultaneously. However, the additional model complexity resulting from the “integrated” strategy can restrict its wide application, which explains its limited use in current studies (see Figure 3).

When both non-DL-based and DL-based imputation models are available, the former may be preferred for its ease of implementation and the fact that non-DL-based methods (such as MICE, Xgboost, LightGBM, etc.) can produce good imputation results when coupled with carefully engineered tabular static and temporal data[14]. The ease of implementation is, however, dependent upon restrictive statistical assumptions about data, which are usually difficult to examine and identify in real-world scenarios[91]. The effort required for feature engineering could be tremendous[27], thereby distracting researchers from their primary research objectives. Moreover, other concerns, such as high dimension[55, 58] and low time efficiency[55, 78], pose obstacles for non-DL-based methods (e.g., k-NN, MICE, etc.). In the presence of healthcare data in complex formats, DL-based models may be a good solution, as statistical assumptions and feature engineering are rarely required and they do not suffer from the curse of dimensionality as many non-DL-based methods do. Additionally, since DL-based models can be directly pre-trained and applied to new samples[61], they may reduce computational costs during evaluation.

Clinical practitioners who lack expertise in deep learning may find it difficult to implement DL-based imputation models, so DL-based models that are specified by data type and packaged well are of practical value. Moreover, researchers should also pay attention to fairness in the imputation process, which does not receive adequate attention at this time, as most studies in this review refer to statistical bias, but do not address social bias, or discrimination against certain groups or individuals[1, 13]. Data imputation influenced by such bias may degrade the subsequent analysis and lead to unfair decision-making, thereby causing inequality and burdening the healthcare system.

This study has several limitations. First, the scope of our review was limited to clinical and translational research; however, alternative DL-based imputation techniques may exist in other research fields and merit examination in a future study. Second, we concentrated primarily on data types and their corresponding imputation strategies, while a more in-depth analysis based on specific research topics may yield additional insights. A discussion of healthcare applications that goes deeper by examining each paper's particular research topics may contribute to the development of effective imputation models.

## 5. Conclusions

Our study fills a gap in the existing literature where DL-based missing value imputation methods have not been systematically reviewed and evaluated. We thoroughly reviewed the DL-based imputation models across a variety of healthcare data types and summarized the missingness mechanisms and appropriate models for each data type. In addition, we identified the advantages of DL-based imputation models and the corresponding integrated strategy. Future research may focus on the portability and fairness of DL-based imputation models with applications to healthcare data.

**Table 1** Definitions of models and corresponding imputation mechanisms

| Model | Definition | Imputation Mechanism[1] |
|---|---|---|
| Multi-layer perceptron (MLP) | It consists of at least three layers of neurons (input, hidden, and output), each of which is fed a non-linear activation function to capture the patterns.[92] | 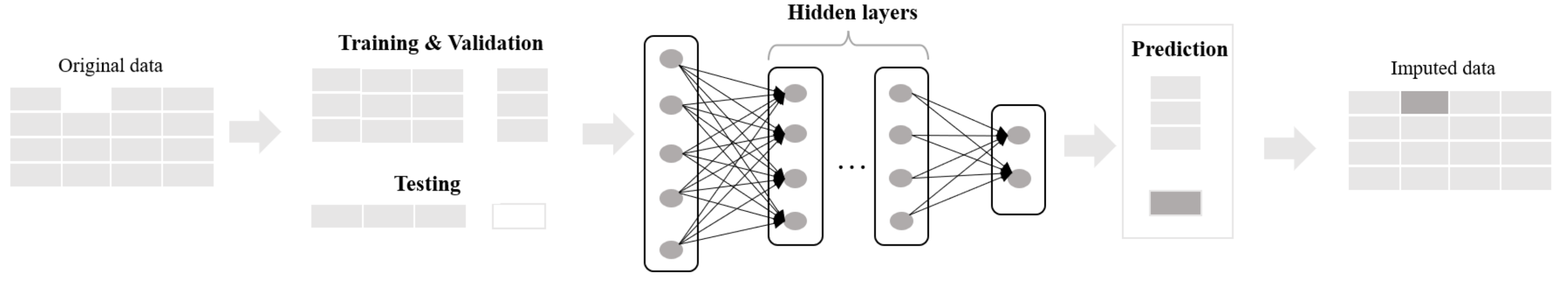 |
| Recurrent neural network (RNN)[2] | It uses the hidden states (memory) to process sequences of inputs. At each step $t$, the input includes the current observation ($x_t$) and the previous hidden state ($h_{t-1}$).[93] | 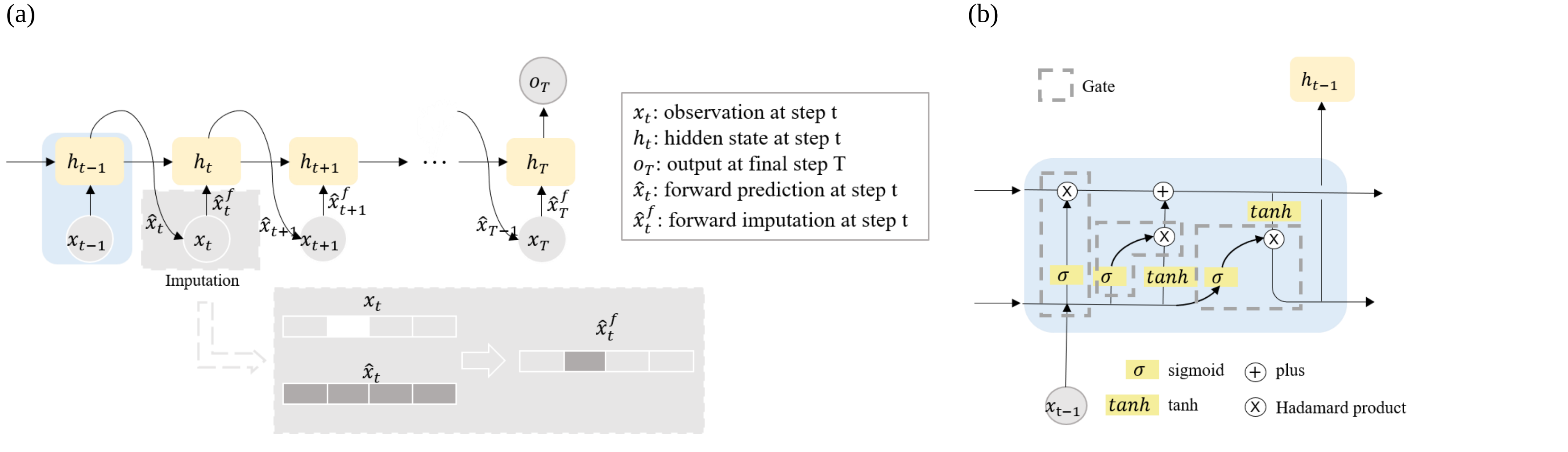 |
| Autoencoder (AE)[3] | It is a framework[4] that co-trains two modules: an encoder that maps the input data into a latent embedding, and a decoder that reconstructs the input from the latent embedding.[94] | 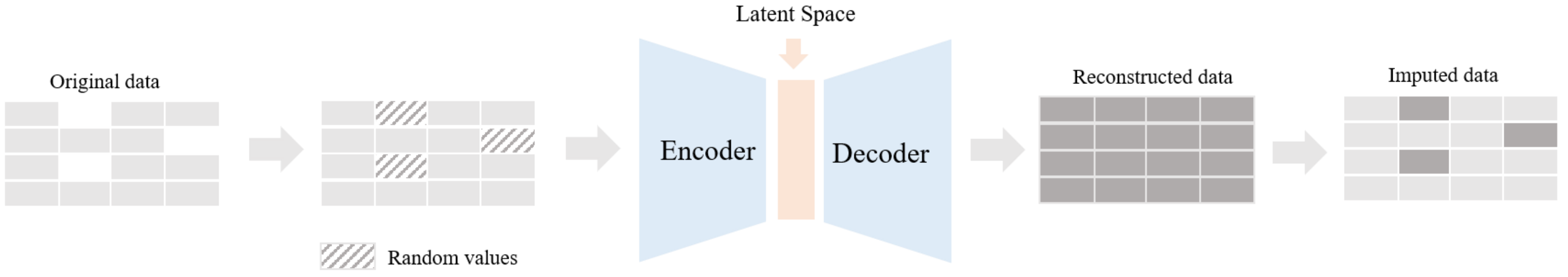 |
| Generative adversarial network (GAN) | It is a framework[4] in an adversarial process, co-training two modules: a generator that captures the data distribution, and a discriminator that detects whether a sample came from the training data rather than the generator.[95] | 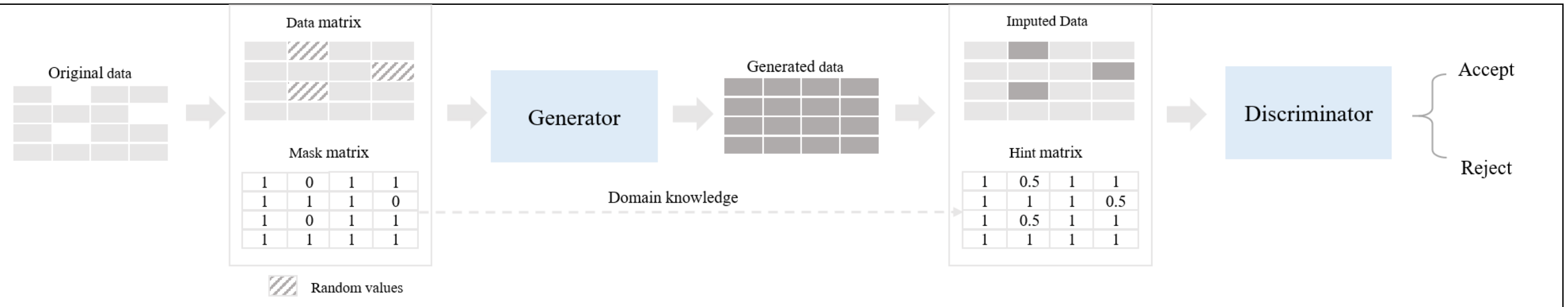 |

[1]*Imputation mechanisms: MLP models can be trained on the complete observations to predict the missing values; RNN models can predict the missing values based on the previous hidden state (forward imputation[45]); Autoencoder can maintain the whole data structure in a good manner and reveal the missing values in its output; GAN can use the generator to capture the data distribution, impute the missing values with the generated data, and apply the discriminator to decide the rightness of the imputation with the assistance of domain knowledge, if applicable. The adversarial process allows for precise data distribution capturing. Based on these general ideas, applications and variants are discussed in subsections 3.1-3.6.*

[2]*Long short-term memory (LSTM) and Gated recurrent unit (GRU) are two main branches of RNN. Compared with vanilla RNN, they have an additional mechanism of "gates" to control the contribution of "memory", i.e., sequence-dependencies, as shown in (b). GRU with two gates is simpler than LSTM with three gates, but performs similarly in many scenarios.*

[3]*Denoise autoencoder (DAE) and variational autoencoder (VAE) hold the same fundamental structure as vanilla autoencoder. DAE receives corrupted data points as input and is trained to predict the uncorrupted data points as its output[96]. Considering missingness as one of the forms of corruption, DAE can be more robust to missing values than vanilla AE. Variational Autoencoder (VAE) utilizes the technique of variational inference in statistics, which introduces probabilistic modeling in latent space to better approximate the true data[61].*

[4]*Framework refers to the fact that the modules (encoder/decoder and generator/discriminator) which respectively shaped the structures of AE and GAN, can embed with various models based on the data input, for example, convolutional neural network (CNN) to tackle images.*

**Figure 1** Preferred Reporting Items for Systematic Reviews (PRISMA) flow diagram

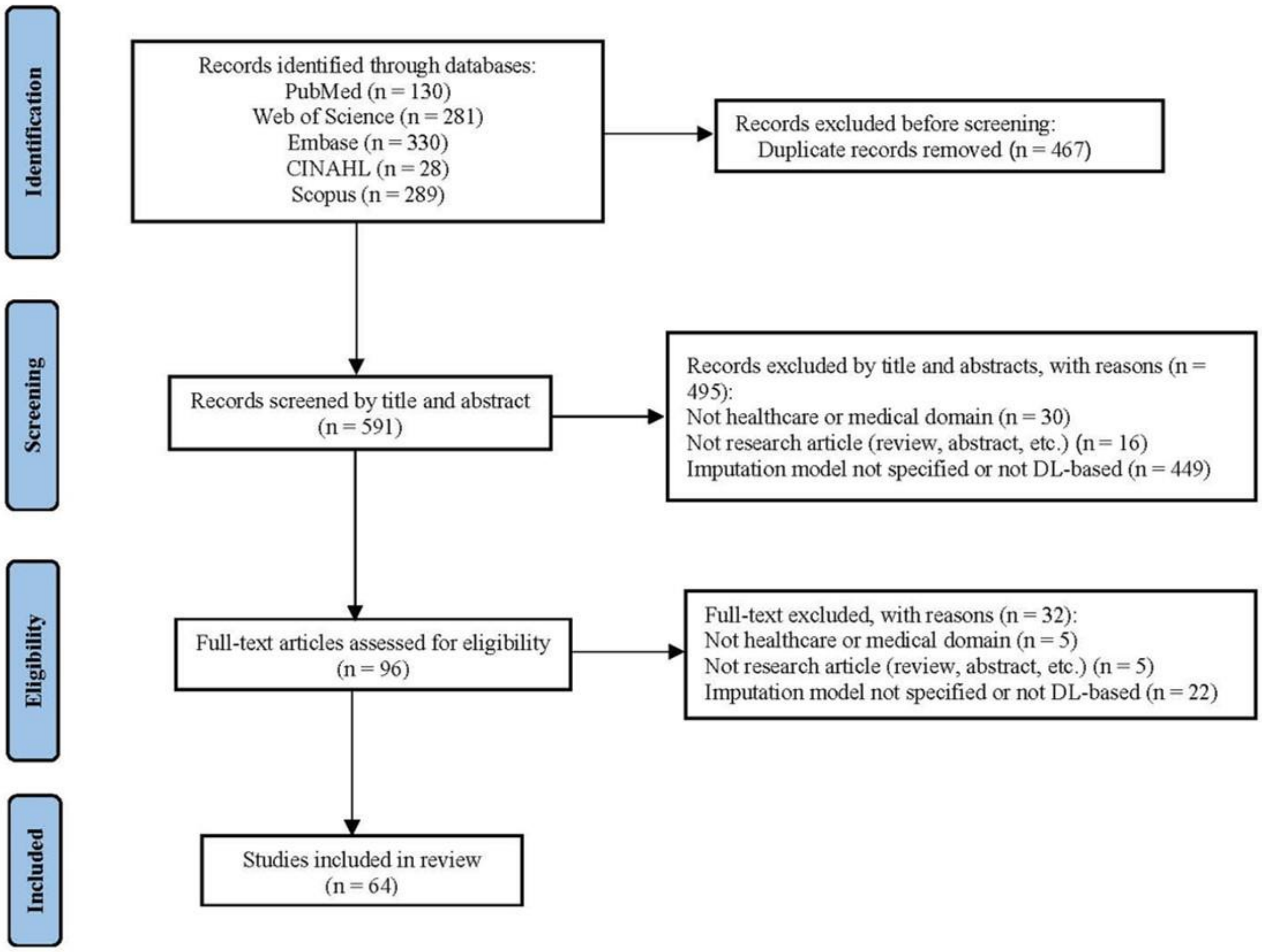

**Figure 2** Evidence map between "backbones" (main architectures) of model and data type

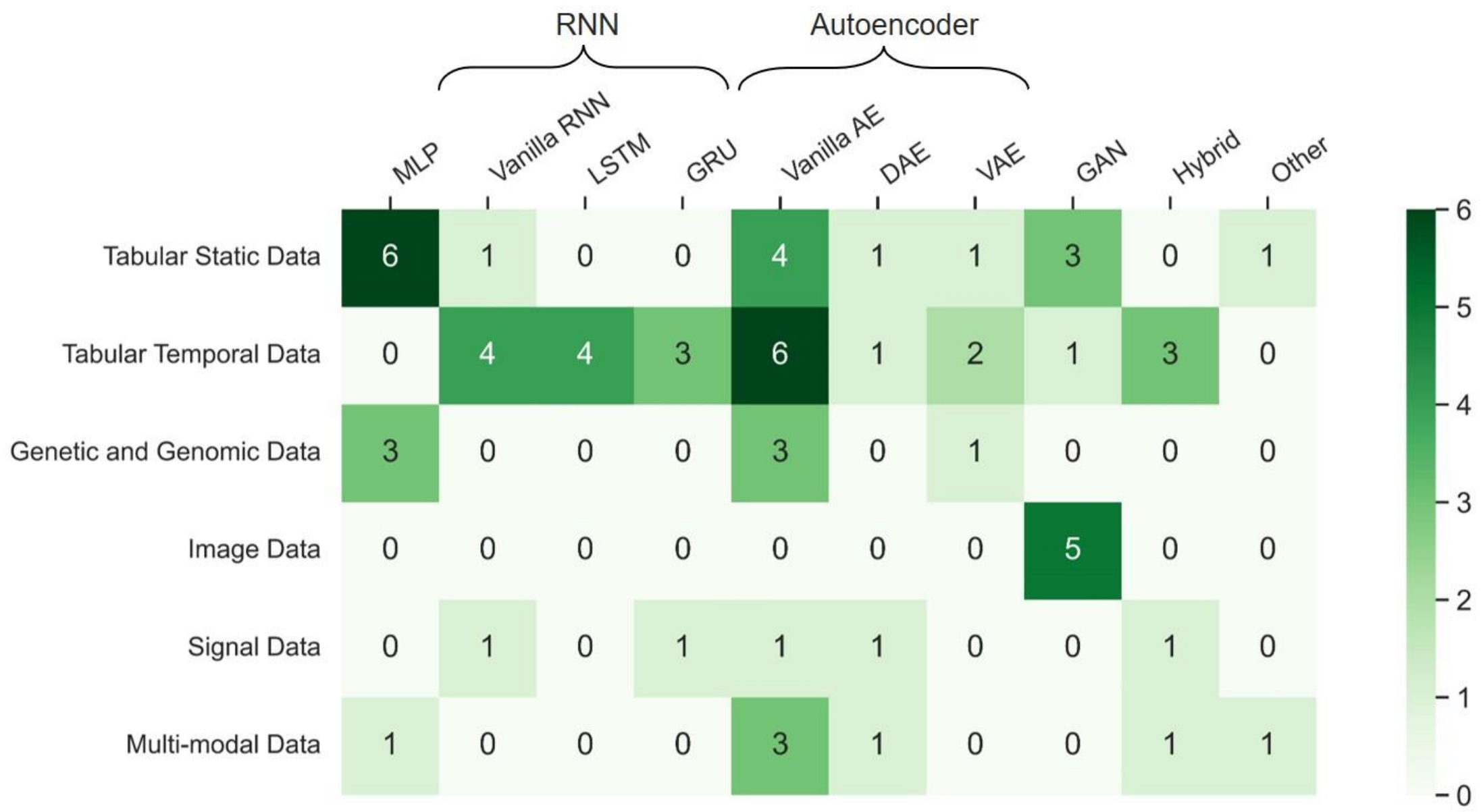

*"Backbones" are classified into ten categories: MLP (multi-layer perceptron), RNN (recurrent neural network), LSTM (Long short-term memory), GRU (gated recurrent unit), AE (autoencoder), DAE (denoising autoencoder), VAE (variational autoencoder), GAN (generative adversarial network) and Other, which includes less frequently used models such as SOM (self-organizing map). Data types are categorized into seven categories: tabular static, tabular temporal, genetic and genomic, image, signal, and multi-modal data. The numbers are non-exclusive.*

**Figure 3** The distribution of imputation frameworks (integrated or separated) by data type

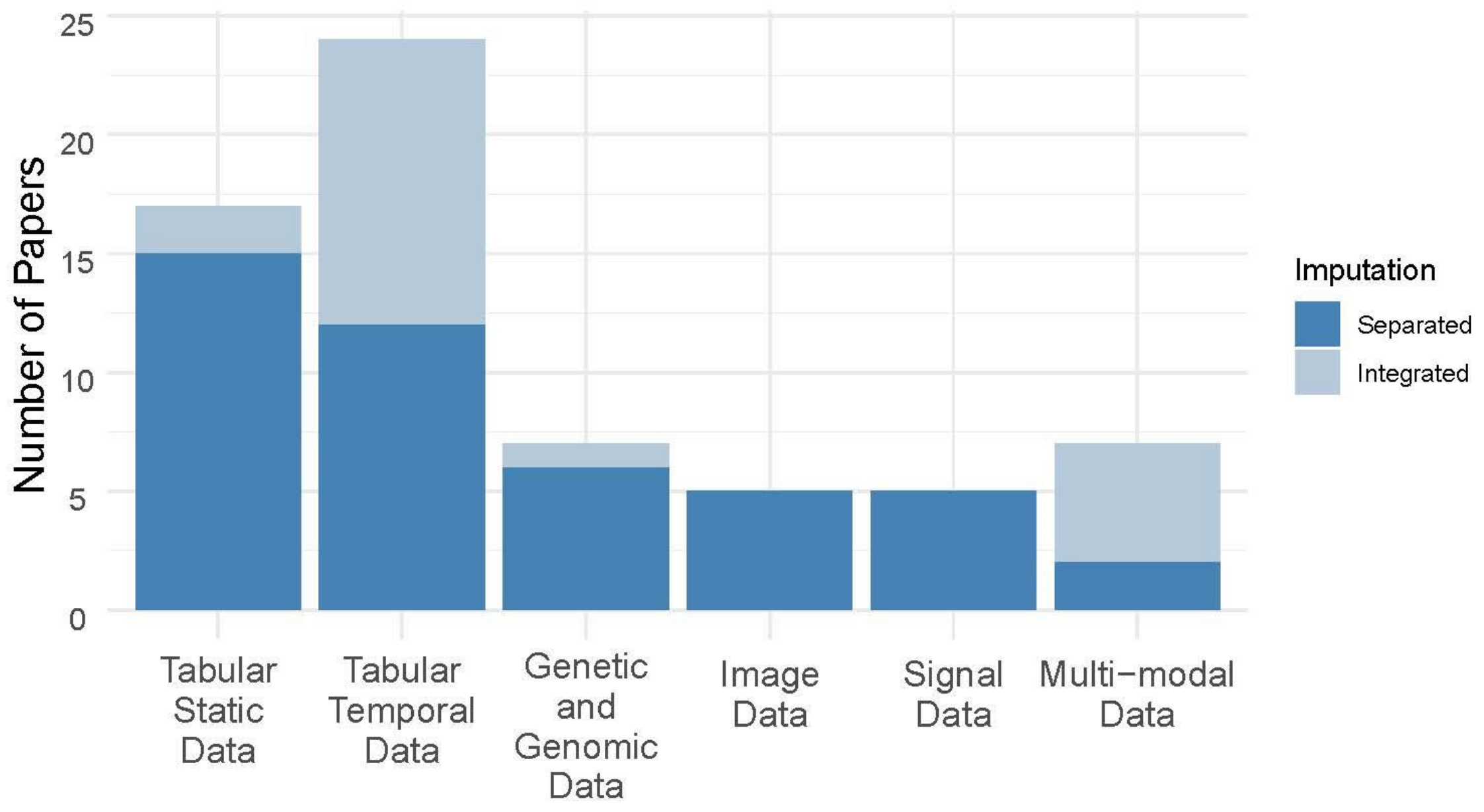

# Supplementary

eTable 1. Search Strategies

| Database | Search Strategy |
|---|---|
| MEDLINE | ("machine learning" OR "deep learning" OR "artificial intelligence" OR "data mining") AND ("missing data" OR "missing value" OR "missing values") AND ("imputation") AND ("medical" OR "clinical" OR "health" OR "healthcare" OR "medicine") |
| Web of Science | ALL=(("machine learning" OR "deep learning" OR "artificial intelligence" OR "data mining")) AND ALL=(("missing data" OR "missing value" OR "missing values")) AND ALL=(("imputation")) AND ALL=(("medical" OR "clinical" OR "health" OR "healthcare" OR "medicine")) |
| Embase | ("machine learning" OR "deep learning" OR "artificial intelligence" OR "data mining") AND ("missing data" OR "missing value" OR "missing values") AND ("imputation") AND ("medical" OR "clinical" OR "health" OR "healthcare" OR "medicine") |
| CINAHL | ("machine learning" OR "deep learning" OR "artificial intelligence" OR "data mining") AND ("missing data" OR "missing value" OR "missing values") AND ("imputation") AND ("medical" OR "clinical" OR "health" OR "healthcare" OR "medicine") |
| Scopus | ("machine learning" OR "deep learning" OR "artificial intelligence" OR "data mining") AND ("missing data" OR "missing value" OR "missing values") AND ("imputation") AND ("medical" OR "clinical" OR "health" OR "healthcare" OR "medicine") |